\documentclass{article}
\usepackage{graphicx}
\usepackage{float}
\usepackage{times}
\usepackage{dsfont}
\usepackage{amsmath,amssymb}
\usepackage{multicol}
\usepackage{siunitx}
\usepackage[bookmarks=true]{hyperref}
\usepackage{cleveref}
\usepackage{xspace}
\usepackage{xurl}
\usepackage{capt-of}
\usepackage[usenames,dvipsnames]{xcolor}
\usepackage{booktabs}  
\usepackage{threeparttable}
\usepackage{multirow}
\usepackage{enumitem}
\usepackage{ulem}
\usepackage{pifont}

\usepackage{wrapfig} 
\usepackage{colortbl}
\usepackage{xcolor}
\usepackage{svg}
\usepackage{subcaption}
\usepackage{siunitx}

\usepackage[preprint]{corl_2025} 

\title{FALCON: Learning Force-Adaptive \\ Humanoid Loco-Manipulation}

\author{
  Yuanhang~Zhang$^{\textnormal{1}}$, Yifu~Yuan$^{\textnormal{1}}$, Prajwal~Gurunath$^{\textnormal{1}}$, 
  Ishita~Gupta$^{\textnormal{1}}$,
  Shayegan~Omidshafiei$^{\textnormal{2}}$\\ 
  \textbf{Ali-akbar~Agha-mohammadi$^{\textnormal{2}}$, Marcell~Vazquez-Chanlatte$^{\textnormal{3}}$, Liam~Pedersen$^{\textnormal{3}}$}\\
  \textbf{Tairan~He$^{\textnormal{1}}$, Guanya~Shi$^{\textnormal{1}}$}\\
  $^{1}$Carnegie Mellon University \quad $^{2}$Field AI \quad $^{3}$Nissan USA \\
  Page: \href{https://lecar-lab.github.io/falcon-humanoid}{\texttt{https://lecar-lab.github.io/falcon-humanoid}} \\ 
  Code: \href{https://github.com/LeCAR-Lab/FALCON}{\texttt{https://github.com/LeCAR-Lab/FALCON}}
  \vspace{-2mm}
}
\usepackage{xcolor}
\newcommand{\goodnumber}[1]{{\color{Methodred}\textbf{#1}}}
\definecolor{Methodred}{RGB}{191, 3, 3}
\definecolor{mydarkblue}{rgb}{0,0.08,0.45}
\definecolor{mydarkgreen}{RGB}{0, 139, 69}
\definecolor{mygreen2}{RGB}{0 205 0}
\definecolor{mybrown}{RGB}{139 69 19}
\definecolor{deepred}{RGB}{200, 30, 30}
\definecolor{deepyellow}{RGB}{230, 180,, 0}
\definecolor{boxblue}{RGB}{79,173,234}
\definecolor{boxgreen}{RGB}{159,206,99}
\hypersetup{
	colorlinks=true,
	linkcolor=blue,
	urlcolor=magenta,
	citecolor=mygreen2,
}

\newcommand{\method}{{\texttt{FALCON}}\xspace}
\newcommand{\decouple}{{Lower-RL-Upper-IK}\xspace}
\newcommand{\wbc}{{Monolithic-Whole-body-RL}\xspace}

\newcommand{\bs}[1]{\boldsymbol{#1}}

\newcommand{\dofposhist}{{\bs{{q}}_{t-4:t}}}
\newcommand{\rootangvelhist}{{\bs{\omega}^{\text{root}}_{t-4:t}}}
\newcommand{\gravityhist}{{\bs{g}_{t-4:t}}}

\newcommand{\dofvelhist}{{\bs{\dot{q}}_{t-4:t}}}

\newcommand{\actionhist}{{\bs{a}_{t-5:t-1}}}

\begin{document}
\maketitle


\begin{figure*}[htb]
    \centering
    \vspace{-6mm}
    \includegraphics[width=\textwidth]{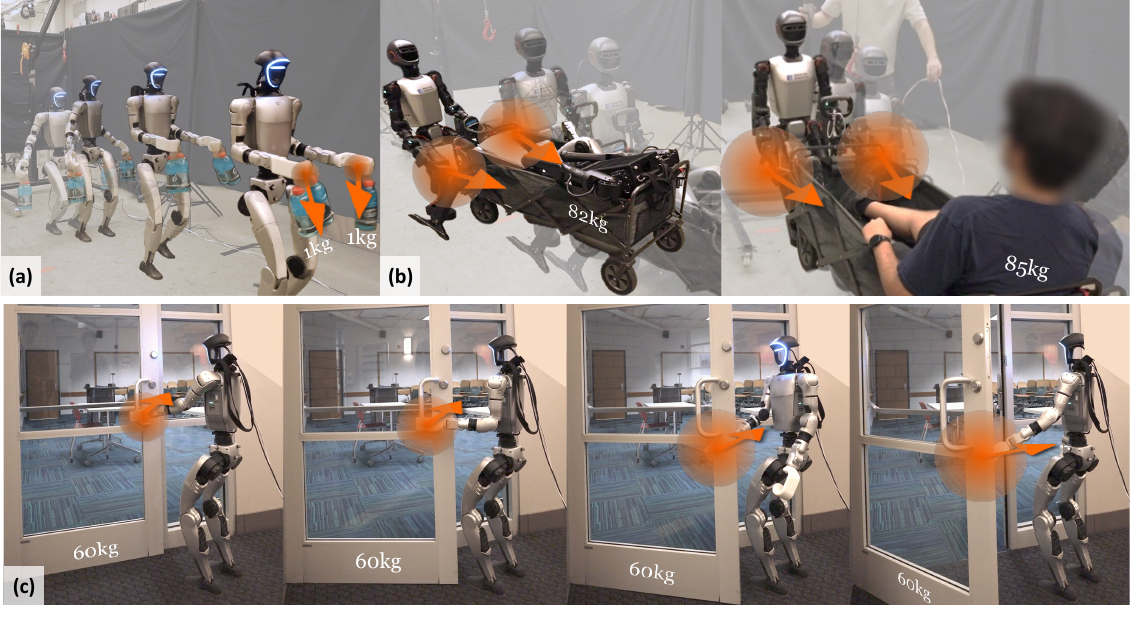}
    \vspace{-6mm}
    \caption{\small \method enables versatile forceful loco-manipulation tasks for humanoids: (a) \textbf{Transporting Payloads}: walk, squat, twist torso with payloads; (b) \textbf{Cart-Pulling} with significant longitudinal forces; (c) \textbf{Door-Opening} using both arms with multi-directional forces. \textbf{Videos}: \href{https://lecar-lab.github.io/falcon-humanoid}{https://lecar-lab.github.io/falcon-humanoid}}
    \label{fig:teaser}
    \vspace{-2.5mm}
\end{figure*}

\begin{abstract}
Humanoid loco-manipulation holds transformative potential for daily service and industrial tasks, yet achieving precise, robust whole-body control with 3D end-effector force interaction remains a major challenge. Prior approaches are often limited to lightweight tasks or quadrupedal/wheeled platforms. To overcome these limitations, we propose \method, a dual-agent reinforcement-learning-based framework for robust force-adaptive humanoid loco-manipulation. \method decomposes whole-body control into two specialized agents: (1) a lower-body agent ensuring stable locomotion under external force disturbances, and (2) an upper-body agent precisely tracking end-effector positions with implicit adaptive force compensation. These two agents are jointly trained in simulation with a force curriculum that progressively escalates the magnitude of external force exerted on the end effector while respecting torque limits. Experiments demonstrate that, compared to the baselines, \method achieves 2$\times$ more accurate upper-body joint tracking, while maintaining robust locomotion under force disturbances and achieving faster training convergence. Moreover, \method enables policy training without embodiment-specific reward or curriculum tuning. Using the same training setup, we obtain policies that are deployed across multiple humanoids, enabling forceful loco-manipulation tasks such as transporting payloads (0-20N force), cart-pulling (0-100N), and door-opening (0-40N) in the real world.

\end{abstract}
\keywords{Humanoid Loco-Manipulation, Force Adaptation, RL} 




\section{Introduction}
\label{sec:introduction}


Humanoid robots have demonstrated remarkable progress in locomotion and manipulation~\cite{gu2025humanoid, darvish2023teleoperation, atlas, unitreeg1, figureai2024, Booster2025T1}. However, extending these capabilities to forceful loco-manipulation remains fundamentally challenging. Tasks such as door opening,
highlighted in the 2015 DARPA Challenge~\cite{dappa}, require not only precise manipulation under dynamic, multi-directional forces but also maintaining lower-body stability throughout the interaction. 
Meeting these demands calls for humanoid systems that can flexibly adapt to varying payloads and contact forces without compromising overall precision and robustness in loco-manipulation.


Reinforcement Learning (RL) has achieved impressive results for humanoid whole-body control~\cite{lu2024mobile, homie, ji2024exbody2, cheng2024expressive, sombolestan2023hierarchical, sombolestan2024adaptive, fey2025bridging, he2024hover, he2024learning, he2024omnih2o, he2025asap, zhuang2025embrace, dantec2021whole, li2023dynamic, murooka2021humanoid, bouyarmane2018quadratic, di2018dynamic, kajita2003biped, fu2024humanplus, zhang2024wococo}, yet existing RL approaches succeed mostly on lightweight tasks but do not consider significant interaction force during loco-manipulation tasks. Currently, there are two main paradigms:
(1) \textit{\decouple}, which applies RL to lower-body locomotion while using kinematic solvers for upper-body control~\cite{lu2024mobile, homie}, lacks whole-body dynamics modeling for forceful interaction and has limited whole-body coordination;
(2) \textit{\wbc}, which directly learns to control all degrees of freedom~\cite{ji2024exbody2, cheng2024expressive}, suffers from inefficient exploration as a single policy must simultaneously learn weakly correlated locomotion and manipulation skills. Although some advances have been made in force adaptation for quadrupeds~\cite{sombolestan2023hierarchical, sombolestan2024adaptive, fey2025bridging, cheng2025rambo, zhong2025bridging}, humanoids pose extra challenges like instability, higher complexity, and stricter torque limits, especially in certain joint configurations.

In this work, we aim to develop an RL framework that enables humanoid robots to perform a diverse set of force-adaptive loco-manipulation tasks. To this end, we introduce \method, a dual-agent RL architecture trained with a carefully designed 3D force curriculum respecting joint torque limits. Our key innovations include:
(1) A \textit{dual-agent learning decomposition} that separates lower-body and upper-body policy training with tailored rewards while sharing the same whole-body proprioception and commands;
(2) A \textit{3D force curriculum} with joint torque feasibility that progressively scales applied 3D forces on both end-effectors while enforcing joint torque constraints through inverse dynamics. \method enables efficient joint training of both stable locomotion and accurate EE tracking in forceful loco-manipulation tasks. We validate \method on Unitree G1 and Booster T1 humanoids, demonstrating its generalization across different platforms through: (1) Transporting Payloads, (2) Cart-Pulling, and (3) Door-Opening (Figure.~\ref{fig:teaser}), which require real-time adaptation to significant unknown 3D interaction force.
In summary, our main contributions are:
\par 
\begin{itemize}[leftmargin=*]
    \item We introduce \method, a dual-agent reinforcement learning framework that enables humanoids to perform forceful loco-manipulation while adapting to substantial, unknown end-effector forces (0–100N, up to 30\% of body weight). \method improves the upper-body joint tracking accuracy over prior methods by 100\% while maintaining robust locomotion performance.
    \item To facilitate the efficient RL training, we design a 3D force curriculum with progressive force application while ensuring joint torque feasibility and maximizing its force-adaptive capability.
    \item We validate \method on two different humanoid platforms (Unitree G1, Booster T1), achieving strong cross-platform generalization with minimal tuning overhead.
\end{itemize}

\section{Related Works}
\label{sec:relatedwork}

\subsection{Humanoid Loco-Manipulation}
Humanoid loco-manipulation remains a challenging control problem in robotics. While traditional model-based methods (e.g., simplified dynamics models and MPC)~\cite{dantec2021whole, sombolestan2024adaptive, li2023dynamic, murooka2021humanoid, bouyarmane2018quadratic, di2018dynamic, kajita2003biped, xue2024full} offer real-time planning, their reliance on manual design limits flexibility and generalizability. In contrast, learning-based methods—particularly sim-to-real RL—have demonstrated promising results in versatile loco-manipulation tasks~\cite{zhang2024catchit, he2024omnih2o, cheng2024expressive, lu2024mobile, homie, dao2024sim, liu2024opt2skill}. For humanoids, two primary paradigms have emerged:
\textit{\decouple} and \textit{\wbc}. For \textit{\decouple}, \citet{lu2024mobile} introduce PMP, which uses inverse kinematics (IK) and PD control for upper body control while locomotion is trained and conditioned on a Conditional Variational Autoencoder (CVAE) representing upper-body motions. Then, \citet{homie} propose HOMIE that follows the same decoupling framework but introduces an exoskeleton-based cockpit for more intuitive human teleoperation. For \textit{\wbc}, \citet{dao2024sim} adopted a unified RL approach for box pick-and-place tasks, training distinct skills (e.g., lifting, walking, stance) and orchestrating them via a finite state machine. \citet{he2024learning, he2024omnih2o} and \citet{ji2024exbody2} employ a teacher-student training framework to mimic human motions for loco-manipulation tasks. 

Despite these advances, few RL methods address significant unknown force disturbances on the EEs for humanoid loco-manipulation, and both paradigms exhibit critical shortcomings accordingly. \textit{\decouple} approaches suffer from delayed force compensation for upper-body control. \textit{\wbc} methods face sample inefficiency from coarsely related task objectives between upper-body manipulation and lower-body locomotion, often leading to overfitting and the behavioral dominance of either upper or lower body. In this study, inspired by~\cite{gronauer2022multi, zhang2021multi, wang2024learning}, we propose \method, a dual-agent RL framework employing task-specific reward formulations for upper-lower body decomposition. Unlike separately trained architectures, the two agents in \method are jointly trained with shared proprioception and commands, allowing mutual awareness of each other's behaviors. This joint training prevents the agents from adapting in isolation and enables coordinated responses to external forces that affect the full-body dynamics.

\subsection{Forceful Interaction in Legged Robots}
Forceful interaction has been extensively studied for quadrupedal robots with mounted arms, through model-based approaches—particularly MPC combined with force planning and control for robust and adaptive locomotion and manipulation~\cite{sombolestan2023hierarchical, sombolestan2024adaptive, rigo2024hierarchical}. Recent advances in RL have further enhanced adaptability, enabling quadrupeds to learn adaptive and agile force interactions including impedance control~\cite{portela2024learning} and aggressive force adaptation~\cite{fey2025bridging}. For humanoids, forceful interaction presents significantly greater challenges due to their more complex dynamics and stringent joint limits. Unlike quadrupeds with centralized mass distributions, humanoids exhibit coupled dynamics between their upper and lower bodies, making force adaptation particularly difficult. Recent model-based approaches have demonstrated force control for heavy-duty tasks~\cite{li2023kinodynamics, murooka2021humanoid}, but these require prior knowledge of manipulated objects' mass, center of mass (CoM), or pre-defined force trajectories, limiting their applicability to unknown disturbances. While some works have attempted explicit force estimation for humanoids~\cite{mattioli2016interaction}, they are restricted to quasi-static scenarios and cannot handle force adaptation in dynamic loco-manipulation scenarios.

In this paper, \method learns to implicitly adapt to unknown external forces on the different EEs with a novel 3D EE force curriculum that considers humanoid joint torque limits. In this way, we can maximize the force adaptability of the learned loco-manipulation policy while ensuring the joint torque limits for robust and safe real-world deployment.

\section{\method: Force-Adaptive Humanoid Loco-Manipulation}
\label{sec:falcon}
\begin{figure*}[tb]
    \centering
    \includegraphics[width=\textwidth]{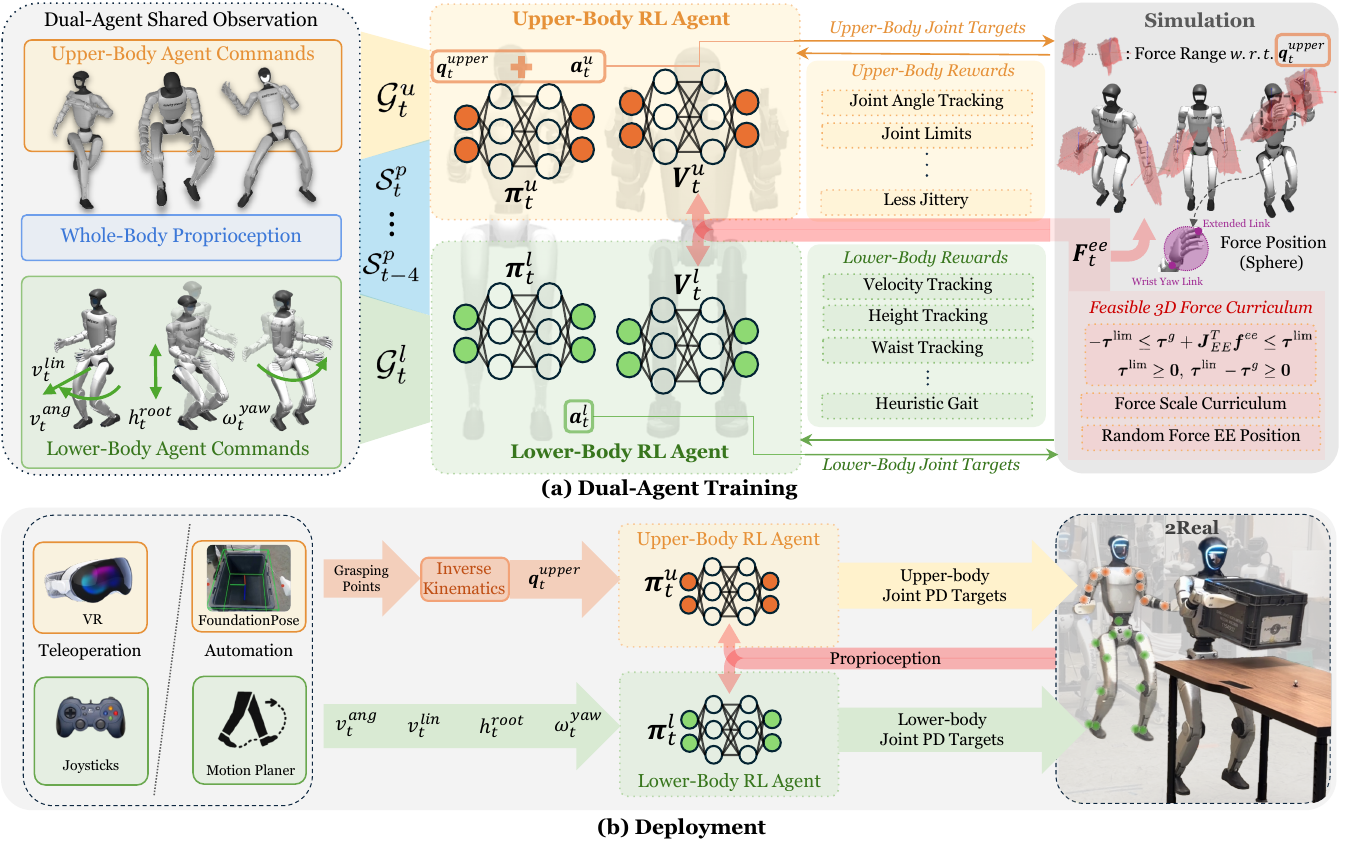}
    \vspace{-4mm}
    \caption{Overview of \method. (a) Two agents with different sub-tasks are jointly trained with shared whole-body proprioception. During training, we apply 3D external forces bounded by upper-body joint torque limits on the end-effectors;
    (b) \method is deployed with either teleoperation or an autonomy pipeline including FoundationPose~\cite{wen2024foundationpose} for pose estimation and motion planning.} 
    \label{fig:falcon}
    \vspace{-4mm}
\end{figure*}


Humanoid loco-manipulation under external EE forces requires coordinated control of both the lower and upper body. We first formulate the problem as a unified \textit{dual goal-conditioned} policy learning problem. Let the degrees of freedom (DoFs) of the humanoid be partitioned into lower-body joints and upper-body joints, with $n^l$ denoting the number of lower-body DoFs, $n^u$ the number of upper-body DoFs, and $n = n^l + n^u$ the total number of actuated joints. 

The robot proprioception $\bs{s}^p_t \in \mathcal{S}_t$ is defined as $\bs{s}_t^{p} \triangleq [\dofposhist, \dofvelhist, \rootangvelhist, \gravityhist, \actionhist]$, which contains five-step histories of joint positions $\bs{q}_t \in \mathbb{R}^{n}$, joint velocities $\bs{\dot{q}}_t \in \mathbb{R}^{n}$, root angular velocity $\bs{\omega}^{\mathrm{root}}_t \in \mathbb{R}^3$, projected gravity $\bs{g}_t \in \mathbb{R}^3$, and previous actions $\bs{a}_{t-1} \in \mathbb{R}^{n}$. The goal space $\mathcal{G}_t$ consists of locomotion goals $\mathcal{G}^l_t \triangleq [\mathbf{v}^{\text{lin,ang}}_t, \phi^{\text{stance}}_t, h^{\text{root}}_t, w^{\text{yaw}}_t]$, specifying desired root linear and angular velocities, stance indicators, root heights, and waist yaw angles, and manipulation goals $\mathcal{G}^u_t \triangleq [\mathbf{q}^{\text{upper*}}_t]$, specifying target joint configurations for the upper body where $\mathbf{q}^{\text{upper*}}_t \in \mathbb{R}^{n^u}$. Under this unified formalism, conventional methods differ mainly in how they generate the action $\bs{a}_t \in \mathbb{R}^{n}$ that commands the robot joints:
\begin{itemize}[leftmargin=*]
    \item \textit{\decouple}: lower-body actions $\bs{a}^l_t \in \mathcal{A}^l_t \subset \mathbb{R}^{n^l}$ are generated by a policy $\pi^l: \bs{s}^p_t \times \langle \mathcal{G}^l_t, \mathcal{G}^u_t \rangle \mapsto \mathcal{A}^l_t$ conditioned on whole-body proprioception and goals, while upper-body actions $\bs{a}^u_t \in \mathcal{A}^u_t \subset \mathbb{R}^{n^u}$ are computed through inverse kinematics (IK) solvers based on $\mathcal{G}^u_t$.
    
    \item \textit{\wbc}: a single policy $\pi: \bs{s}^p_t \times \mathcal{G}_t \mapsto \bs{a}_t$ directly predicts the full-body action $\bs{a}_t \in \mathbb{R}^{n}$, attempting to satisfy both locomotion and manipulation objectives simultaneously.
\end{itemize}

While \textit{\decouple} methods are sample-efficient, they neglect upper-body force compensation and whole-body coupling under EE force disturbances. In contrast, \textit{\wbc} methods improve expressiveness but suffer from exploration inefficiency due to the large action space spanning coarsely related locomotion and manipulation objectives. To overcome these challenges, we introduce \method, a dual-agent RL framework that achieves training efficiency and coordination through decomposition learning with shared whole-body observation.

\subsection{Dual-Agent Learning Framework}
\label{subsec:dual-agent}

As shown in \Cref{fig:falcon}, \method jointly trains two agents, each specialized for a different subtask. The lower-body locomotion agent learns a policy $\pi^l: \bs{s}^p_t \times \mathcal{G}^l_t \mapsto \mathcal{A}^l_t$ with value function $V^l(\cdot)$, while the upper-body manipulation agent learns a policy $\pi^u: \bs{s}^p_t \times \mathcal{G}^u_t \mapsto \mathcal{A}^u_t$ with value function $V^u(\cdot)$. Both agents observe the same proprioceptive input $\bs{s}^p_t$ but optimize independent goal-conditioned objectives:
\begin{align}
    r^l_t = \mathcal{R}^l(\bs{s}^p_t, \mathcal{G}^l_t)~\text{(locomotion)} \quad
    r^u_t = \mathcal{R}^u(\bs{s}^p_t, \mathcal{G}^u_t)~\text{(manipulation)}
\end{align}
These two policy parameters $\theta_l$ and $\theta_u$ are updated via proximal policy optimization (PPO~\cite{PPO}):
\begin{align}
    \max_{\theta_l} \mathbb{E} \left[\sum_{t=1}^T \gamma^{t-1} r^l_t\right]~\text{(Lower-body)} \quad
    \max_{\theta_u} \mathbb{E} \left[\sum_{t=1}^T \gamma^{t-1} r^u_t\right]~\text{(Upper-body)}
\end{align}

where $\gamma$ is the discount factor. The upper-body target joint angles $\mathbf{q}^{\text{upper}}_t$ (target joints of shoulders, elbows, wrists) are randomly sampled from the AMASS dataset~\cite{AMASS} during training, and calculated via IK during deployment. The combined action from the two agents $\bs{a}_t = [\bs{a}^l_t; \bs{a}^u_t]$ is sent to a joint-level PD controller. As the real-world humanoid control is inherently partially observable, we adopt asymmetric actor-critic training, where critics additionally access privileged information including root linear velocities and EE forces $\bs{F}^{ee}_t$ during training but not during deployment. Detailed reward designs and domain randomization during training are provided in \Cref{appendix:reward} and \Cref{appendix:dr}.


\subsection{Torque-Limit-Aware 3D Force Curriculum}
\label{subsec:force_curriculum}
For humanoid robots—particularly those with relatively weak joint torque limits, such as the wrist joints on the Unitree Humanoid G1— it is crucial to explicitly account for these torque constraints when large external disturbances are applied to the end-effectors (EEs). Ignoring these limits during upper body policy training can lead to
unexpected or unsafe behaviors due to torque saturation or joint limit violations in real-robot deployment. Additionally, it's important to gradually increase the external force during training, allowing the policy to progressively learn effective force adaptation strategies. To achieve these, our force application framework follows through three principles:

\paragraph{Torque-Aware Force Computation:}
Before applying forces, we first need to estimate the maximum forces that we can exert on the left or right end-effector. Given the left or right end-effector Jacobian $\bs{J}_{EE} \in \mathbb{R}^{3\times \frac{n_u}{2}}$ at its Center of Mass (CoM), their joint torque limit $\bs{\tau}^{\lim} \in \mathbb{R}^{3\times\frac{n_u}{2}}$ (with $\bs{\tau}^{\lim} \geq \bs{0}$), and the gravity compensation torque $\bs{\tau}^g \in \mathbb{R}^{3\times\frac{n_u}{2}}$ which satisfies $-\bs{\tau}^{\lim} \leq \bs{\tau}^g \leq \bs{\tau}^{\lim}$ to ensure feasibility, we estimate the maximum and minimum admissible forces $\bs{f}^{\max}, \bs{f}^{\min}$ along each Cartesian axis $i \in {x, y, z}$ by analyzing the worst-case joint torque induced by a unit force applied in each direction. The element-wise force bound can be computed in parallel as:
\begin{equation}
-\bs{\tau}^{\lim} \leq \bs{\tau}^g + \bs{J}_{EE}^T \bs{f}^{ee} \leq \bs{\tau}^{\lim}
\end{equation}
\begin{equation}
f_i^{\max} = \min_j \left( \frac{\tau_j^{\lim} - \tau_j^g}{|J_{EE}^{ji}| + \epsilon} \right), \quad
f_i^{\min} = \max_j \left( \frac{-\tau_j^{\lim} - \tau_j^g}{|J_{EE}^{ji}| + \epsilon} \right), 
\end{equation}

where $J_{EE}^{ji}$ denotes the $(j,i)$-th element of the end-effector Jacobian matrix, and $\epsilon$ is a small positive constant to prevent division by zero. After that, we sample the relatively ratio $\bs{\gamma} = [\gamma_x, \gamma_y, \gamma_z]$ among x, y and z axis through Dirichlet Distribution \cite{ng2011dirichlet}, which satisfy $\sum_{i \in \{x,y,z\}} \gamma_i=1$. The feasible applied force will be uniformly sampled within the estimated range and expressed as: 
\begin{equation}
\bs{f}^{ee}_t = \sum_{i \in \{x,y,z\}} F_i \cdot \bs{e}_i, \quad \text{where} \quad F_i \sim \mathcal{U}[\gamma_i \cdot f_i^{\min}, \gamma_i \cdot f_i^{\max}]
\end{equation}

This approach maximizes force adaptivity while respecting torque limits, leading to more effective training than random sampling, as explained in \Cref{appendix:force_curr}. Note that applied forces may differ between left and right EEs due to asymmetric upper-body configurations (\Cref{fig:falcon}).

\paragraph{Progressive Force Curriculum:}
To facilitate progressive force adaptation, the estimated EE forces are scaled by a global factor \(\alpha_g \in (0,1)\), increasing over training, so the applied force becomes \(\bs{F}^{ee}_t = \alpha_g \cdot \bs{f}_t^{ee}\). During walking, planar forces are projected opposite to the velocity. A low-pass filter is applied to reduce force jitter.

\paragraph{Position Randomization of the Applied Force:}
Learning-based force adaptation leverages proprioceptive history to implicitly compensate for external forces, removing the need for explicit force estimation~\cite{mattioli2016interaction} or sensing~\cite{guo2024flying}. To improve robustness to variations in end-effector (EE) contact points—which alter the torque mapping via the EE Jacobian—we randomize force application along the EE link, from the wrist yaw to the distal segment, as illustrated in \Cref{fig:falcon}. 


\section{Simulation and Real-World Experiments}
\label{sec:experiments}
In this section, we present extensively quantitative comparison between \method and the baselines as well as qualitative results on real-world deployment. We choose Unitree Humanoid G1 and Booster T1 as our humanoid platforms. Specifically, we address the following key questions:

\textbf{Q1}: Can \method outperform other baselines in terms of both upper-body manipulation and lower-body locomotion performance? \par
\textbf{Q2}: Why does \method has better training-efficiency compared to \wbc (M-WB-RL) for force-adaptive loco-manipulation? \par
\textbf{Q3}: Does \method work for different humanoids to show cross-platform generalizability? 

\begin{table*}[t]
\centering
\resizebox{\linewidth}{!}{%
\begingroup
\setlength{\tabcolsep}{4pt}
\renewcommand{\arraystretch}{1.2}
\begin{tabular}{llrrrrrr}
\toprule
\multicolumn{2}{c}{} & \multicolumn{3}{c}{$E_{\text{tracking}}^{\text{upper}}\downarrow$} & \multicolumn{3}{c}{$E_{\text{tracking}}^{\text{root}}\downarrow$} 
\\ 
\cmidrule(r){1-2} \cmidrule(r){3-5} \cmidrule(r){6-8}
\multicolumn{2}{c}{Methods} & $\text{N-Force}$ & $\text{M-Force}$ & $\text{L-Force}$ & $\text{N-Force}$ & $\text{M-Force}$ & $\text{L-Force}$ \\
\cmidrule(r){1-2} \cmidrule(r){3-5} \cmidrule(r){6-8}
\multirow{4}{*}{\shortstack{Lower-RL \\ Upper-IK}} 
& PD-w/o-Force-Curr.  & 0.46 \tiny $\pm$ 0.04 & 0.94 \tiny $\pm$ 0.05 & 1.44 \tiny $\pm$ 0.06 & 0.38 \tiny $\pm$ 0.04 & 0.66 \tiny $\pm$ 0.05 & 1.14 \tiny $\pm$ 0.06 \\
& PD-Force-Curr.      & 0.44 \tiny $\pm$ 0.03 & 0.93 \tiny $\pm$ 0.03 & 1.42 \tiny $\pm$ 0.04 & 0.33 \tiny $\pm$ 0.03 & 0.38 \tiny $\pm$ 0.03 & 0.46 \tiny $\pm$ 0.03 \\
& PID-Force-Curr.     & 0.24 \tiny $\pm$ 0.03 & 0.31 \tiny $\pm$ 0.03 & 0.60 \tiny $\pm$ 0.04 & 0.32 \tiny $\pm$ 0.03 & 0.35 \tiny $\pm$ 0.03 & 0.46 \tiny $\pm$ 0.04 \\
& PD-ID-Force-Curr.   & 0.29 \tiny $\pm$ 0.03 & 0.38 \tiny $\pm$ 0.03 & 0.53 \tiny $\pm$ 0.04 & 0.40 \tiny $\pm$ 0.03 & 0.42 \tiny $\pm$ 0.03 & 0.47 \tiny $\pm$ 0.04 \\
\cmidrule(r){1-2} \cmidrule(r){3-5} \cmidrule(r){6-8}
\multirow{2}{*}{M-WB-RL} 
& w/o-Force-Curr.           & 0.46 \tiny $\pm$ 0.05 & 1.03 \tiny $\pm$ 0.07 & 1.65 \tiny $\pm$ 0.08 & 0.34 \tiny $\pm$ 0.04 & 0.67 \tiny $\pm$ 0.05 & 1.28 \tiny $\pm$ 0.07 \\
& with-Force-Curr.          & 0.43 \tiny $\pm$ 0.04 & 0.50 \tiny $\pm$ 0.04 & 0.73 \tiny $\pm$ 0.05 & \goodnumber{0.28} \tiny $\pm$ 0.03 & \goodnumber{0.32} \tiny $\pm$ 0.03 & \goodnumber{0.44} \tiny $\pm$ 0.04 \\
\cmidrule(r){1-2} \cmidrule(r){3-5} \cmidrule(r){6-8}
\multirow{2}{*}{\goodnumber{\method}} 
& w/o-Force-Curr.           & \goodnumber{0.14} \tiny $\pm$ 0.03 & 0.55 \tiny $\pm$ 0.04 & 1.06 \tiny $\pm$ 0.06 & 0.32 \tiny $\pm$ 0.03 & 0.66 \tiny $\pm$ 0.05 & 1.24 \tiny $\pm$ 0.06 \\
& with-Force-Curr.          & 0.21 \tiny $\pm$ 0.03 & \goodnumber{0.24} \tiny $\pm$ 0.03 & \goodnumber{0.37} \tiny $\pm$ 0.04 & \goodnumber{0.27} \tiny $\pm$ 0.03 & \goodnumber{0.30} \tiny $\pm$ 0.03 & \goodnumber{0.45} \tiny $\pm$ 0.04 \\
\bottomrule
\end{tabular}
\endgroup}
\vspace{-2mm}
\caption{Loco-Manipulation Evaluation of \method and Baselines in IsaacGym.}
\label{tab:loco-manip}
\vspace{-4mm}
\end{table*}

\subsection{Evaluation Criterion}
\label{subsec:eval_criterion}
To evaluate the performance of the learned low-body locomotion and upper-body manipulation capabilities, we consider the following metrics under dynamically unknown and 3D EE forces $\textbf{F}_t \in \mathbb{R}^3$, given a sequence of target upper-body joints $\bs{q}^{\text{upper*}}_t$, target root velocities $\bs{v}^{\text{lin,ang*}}_t$ and stance signal $\phi_t^{\text{stance}}$, where $t=1,2,...,T$ and $T$ is the sequence length:\par


\textit{(ii) Upper-Body Joints Tracking Error}:
$
E_{\text{tracking}}^{\text{upper}}(\bs{q}^{\text{upper*}}_t) = \frac{1}{T} \sum_{t=1}^T \left| \bs{q}_t^{\text{upper}} - \bs{q}_t^{\text{upper*}} \right|
$

\textit{(iii) Root Velocity Tracking Error:} 
$
E_{\text{tracking}}^{\text{root}}(\bs{v}_t^{\text{lin,ang*}}) = \frac{1}{T} \sum_{t=1}^T \left| \bs{v}_t^{\text{lin,ang}} - \bs{v}_t^{\text{lin,ang*}} \right|
$
\subsection{Baselines}

We consider two types of baseline methods for force adaptation, both trained under the same goal space (e.g., commands) in \Cref{subsec:dual-agent} and force curriculum described in \Cref{subsec:force_curriculum}, with each type further including relevant ablation variants.


\textbf{Decoupled Lower-body RL with Upper-body IK Controllers.} For all variants, RL is used for lower-body locomotion, and IK provides target upper-body joint angles from end-effector poses. The key differences lie in the use of force curriculum and the upper-body joint tracking strategy:

\begin{enumerate}[leftmargin=*, label=(\alph*)]
    \item \textit{Upper-PD-w/o-Force-Curr.}: A baseline following~\cite{lu2024mobile, homie}, using PD control for upper-body joint tracking without force randomization.
    \item \textit{Upper-PD}: Extends (a) by incorporating force curriculum, enabling lower-body adaptation to external forces; upper-body remains PD-controlled.
    \item \textit{Upper-PID}: Extends (b) by adding an integral term to the upper-body controller to reduce steady-state tracking error.
    \item \textit{Upper-PD-ID}: Extends (a) with a learned force estimator~\cite{portela2024learning} and inverse-dynamics-based torque compensation under quasi-static assumptions (details in \Cref{appendix:force_est}).
\end{enumerate}

\textbf{Monolithic Whole-body RL}

\begin{enumerate}[leftmargin=*, label=(\alph*)]
    \addtocounter{enumi}{4}
    \item \textit{Monolithic-WB-RL-w/o-Force-Curr.}: Built upon prior designs~\cite{cheng2024expressive, he2024hover}, a single agent is trained with the same goal commands as \method, but without applying any force during training.
    \item \textit{Monolithic-WB-RL-with-Force-Curr.}: Based on (e), we adopt force randomization into the training curriculum for force adaptation, while keeping the other training settings identical.
\end{enumerate}

\subsection{Simulation Results}
To answer \textbf{Q1} (\textit{Can \method outperform other baselines in terms of both upper-body manipulation and lower-body locomotion performance?}) and \textbf{Q2} (\textit{Why does \method has better training-efficiency compared to \wbc (M-WB-RL) for force-adaptive loco-manipulation?}), we conduct quantitative comparisons of our method with other two baselines in IsaacGym on Unitree Humanoid G1.

\paragraph{Loco-Manipulation Performance:}
We evaluate \method and baselines on 252 ACCAD~\cite{accad} motion targets under three force levels: (i) No-Force ($\alpha_g=0$), (ii) Middle-Force ($\alpha_g=0.5$), and (iii) Large-Force ($\alpha_g=1.0$), applied to both end-effectors. As shown in \Cref{tab:loco-manip}, across all settings, \method with force curriculum achieves the lowest tracking errors in both upper-body motion ($E^{\text{upper}}_{\text{tracking}}$) and root velocity ($E^{\text{root}}_{\text{tracking}}$), demonstrating robust manipulation under disturbance. Under L-Force, it reduces upper-body error to 0.37, outperforming PID-Force-Curr. (0.60) and M-WB-RL (0.73). Root error remains low at 0.45, indicating stable locomotion. While force curriculum benefits all methods, \method gains most due to its decomposed learning structure.

\begin{wrapfigure}{r}{0.6\textwidth}
    \centering
    \vspace{-0mm}
    \includegraphics[width=\linewidth]{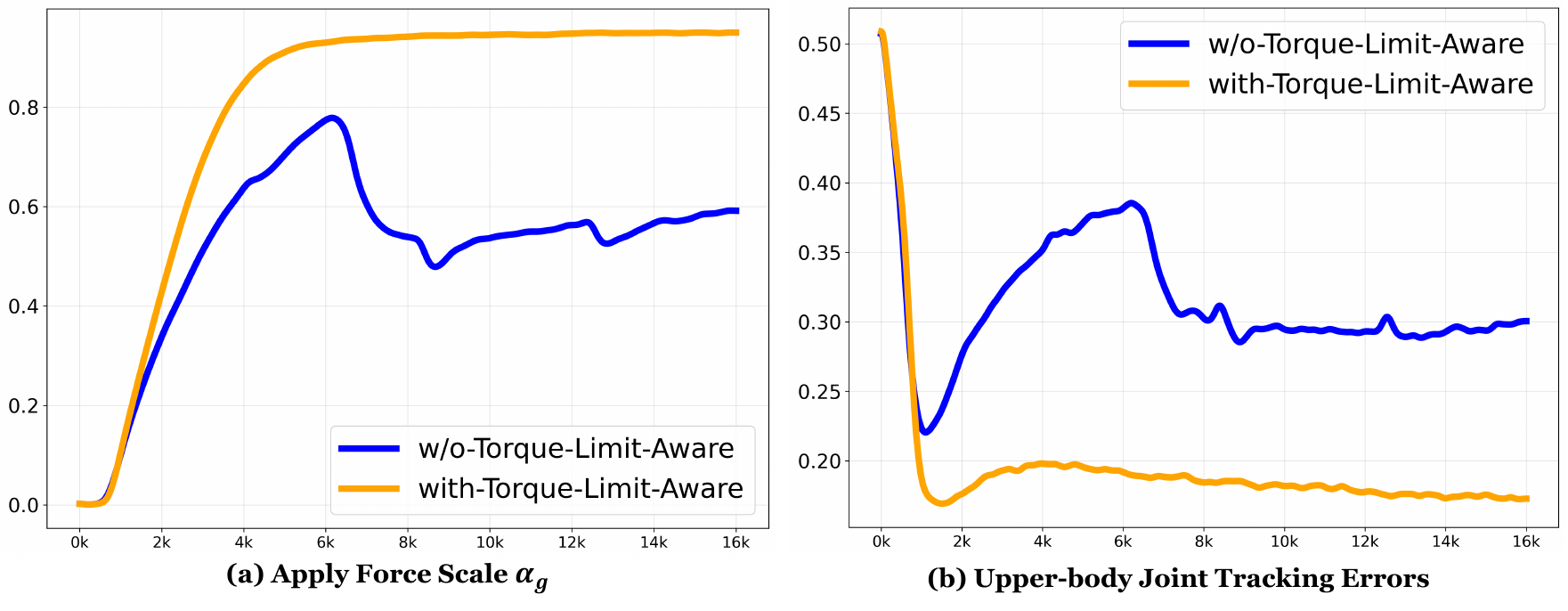}
    \caption{\small{(a) Progression of the Apply Force Scale $\alpha_g$; (b) Upper-body Joint Tracking Errors During Training}}
    \label{fig:force_curr}
    \vspace{-3mm}
\end{wrapfigure}

\paragraph{Torque-Limit-Aware Force Curriculum}
\label{appendix:force_curr}
To assess the effectiveness of the proposed torque-limit-aware force curriculum (\Cref{sec:falcon}), we compare it with a baseline that samples random forces from a wide clipping range ($X: [-100\mathrm{N}, 100\mathrm{N}]$, $Y: [-100\mathrm{N}, 100\mathrm{N}]$, $Z: [-100\mathrm{N}, 5\mathrm{N}]$) without enforcing torque feasibility. Training curves and quantitative results are shown in \Cref{fig:force_curr} and \Cref{tab:force_curr}. During evaluation, applied forces remain bounded by the estimated admissible limits.

\begin{table*}[t]
\centering
\resizebox{\linewidth}{!}{%
\begingroup
\setlength{\tabcolsep}{4pt}
\renewcommand{\arraystretch}{1.2}
\begin{tabular}{llrrrrrr}
\toprule
\multicolumn{2}{c}{} & \multicolumn{3}{c}{$E_{\text{tracking}}^{\text{upper}}\downarrow$} & \multicolumn{3}{c}{$E_{\text{tracking}}^{\text{root}}\downarrow$} 
\\ 
\cmidrule(r){1-2} \cmidrule(r){3-5} \cmidrule(r){6-8}
\multicolumn{2}{c}{Methods} & $\text{N-Force}$ & $\text{M-Force}$ & $\text{L-Force}$ & $\text{N-Force}$ & $\text{M-Force}$ & $\text{L-Force}$ \\
\cmidrule(r){1-2} \cmidrule(r){3-5} \cmidrule(r){6-8}
\multirow{2}{*}{\goodnumber{\method}} 
& w/o-Torque-Limit-Aware.           & 0.42 \tiny $\pm$ 0.07 & 0.45 \tiny $\pm$ 0.08 & 0.61 \tiny $\pm$ 0.10 & 0.45 \tiny $\pm$ 0.03 & 0.46 \tiny $\pm$ 0.05 & 0.54 \tiny $\pm$ 0.05 \\
& with-Torque-Limit-Aware.          & \goodnumber{0.23} \tiny $\pm$ 0.03 & \goodnumber{0.26} \tiny $\pm$ 0.03 & \goodnumber{0.36} \tiny $\pm$ 0.04 & \goodnumber{0.30} \tiny $\pm$ 0.03 & \goodnumber{0.32} \tiny $\pm$ 0.03 & \goodnumber{0.39} \tiny $\pm$ 0.02 \\
\bottomrule
\end{tabular}
\endgroup}
\vspace{-1mm}
\caption{Evaluation of \method using torque-limit-aware (Max-Force-Estimation) curriculum versus \textit{w/o torque-limit-aware} force curriculum in IsaacGym. Our curriculum achieves significantly better tracking performance, especially for upper-body manipulation under large forces.}
\label{tab:force_curr}
\vspace{-4mm}
\end{table*}

\Cref{fig:force_curr}~(a) shows that  force curriculum saturates at a force scale $\alpha_g = 0.6$ due to frequent violations of torque limits, which hinder further progression. Additionally, as illustrated in \Cref{fig:force_curr}~(b), the force curriculum \textit{w/o torque-limit-aware} results in larger upper-body tracking errors, since excessive forces regularly exceed the feasible torque bounds, impairing the learning of effective upper body force compensation. Consequently, as shown in \Cref{tab:force_curr}, policies trained \textit{w/o torque-limit-aware} force curriculum tend to overfit to the locomotion objective, compromising upper-body accuracy. In contrast, our torque-limit-aware curriculum facilitates balanced learning of both upper-body joint tracking and root velocity tracking under significant external disturbances.

Note that in \Cref{tab:loco-manip}, we use a narrower force clipping range ($X: [-50\mathrm{N}, 50\mathrm{N}]$, $Y: [-50\mathrm{N}, 50\mathrm{N}]$, $Z: [-60\mathrm{N}, 5\mathrm{N}]$) compared to the wider range in \Cref{tab:force_curr} ($X: [-100\mathrm{N}, 100\mathrm{N}]$, $Y: [-100\mathrm{N}, 100\mathrm{N}]$, $Z: [-100\mathrm{N}, 5\mathrm{N}]$). The results in \Cref{tab:force_range} show that increasing the force range has minimal impact on loco-manipulation performance, highlighting the robustness of our torque-limit-aware force curriculum.

\begin{table*}[tb]
\centering
\resizebox{\linewidth}{!}{%
\begingroup
\setlength{\tabcolsep}{4pt}
\renewcommand{\arraystretch}{1.2}
\begin{tabular}{llrrrrrr}
\toprule
\multicolumn{2}{c}{} & \multicolumn{3}{c}{$E_{\text{tracking}}^{\text{upper}}\downarrow$} & \multicolumn{3}{c}{$E_{\text{tracking}}^{\text{root}}\downarrow$} 
\\ 
\cmidrule(r){1-2} \cmidrule(r){3-5} \cmidrule(r){6-8}
\multicolumn{2}{c}{Methods} & $\text{N-Force}$ & $\text{M-Force}$ & $\text{L-Force}$ & $\text{N-Force}$ & $\text{M-Force}$ & $\text{L-Force}$ \\
\cmidrule(r){1-2} \cmidrule(r){3-5} \cmidrule(r){6-8}
\multirow{2}{*}{\goodnumber{\method}} 
& Smaller-Force-Clip.          & \goodnumber{0.21} \tiny $\pm$ 0.03 & \goodnumber{0.24} \tiny $\pm$ 0.03 & \goodnumber{0.37} \tiny $\pm$ 0.04 & \goodnumber{0.27} \tiny $\pm$ 0.03 & \goodnumber{0.30} \tiny $\pm$ 0.03 & \goodnumber{0.45} \tiny $\pm$ 0.04 \\
& Larger-Force-Clip.          & \goodnumber{0.23} \tiny $\pm$ 0.03 & \goodnumber{0.26} \tiny $\pm$ 0.03 & \goodnumber{0.36} \tiny $\pm$ 0.04 & \goodnumber{0.30} \tiny $\pm$ 0.03 & \goodnumber{0.32} \tiny $\pm$ 0.03 & \goodnumber{0.39} \tiny $\pm$ 0.02 \\
\bottomrule
\end{tabular}
\endgroup}
\vspace{-1mm}
\caption{Evaluation of \method with a smaller force clip range in \Cref{tab:loco-manip} versus a larger force clip range in \Cref{tab:force_curr}.}
\label{tab:force_range}
\vspace{-0mm}
\end{table*}

\paragraph{Exploration and Learning:}
\begin{wrapfigure}{r}{0.65\linewidth}
    \centering
    \vspace{-6mm}
    \includegraphics[width=1\linewidth]{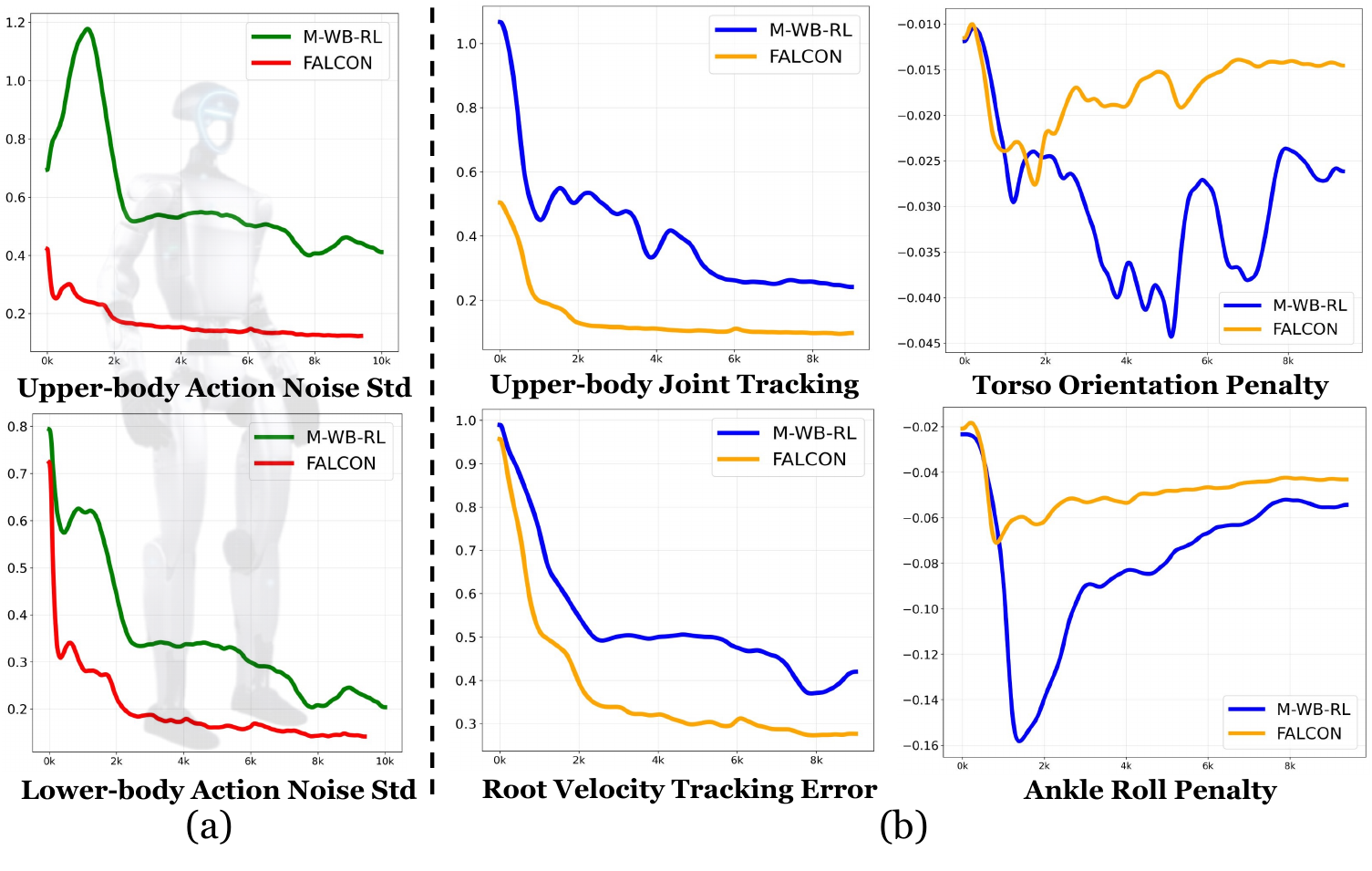}
    \vspace{-7mm}
    \caption{Comparison of FALCON and M-WB-RL: (a) action noise std; (b) tracking errors and penalties.}
    \label{fig:rewards}
    \vspace{-2mm}
\end{wrapfigure}
(i) \textit{Action Noise Std:} As shown in \Cref{fig:rewards}~(a), \method exhibits faster and smoother noise decay in both upper and lower body actions, indicating more efficient and stable exploration. In contrast, M-WB-RL suffers from prolonged noise due to entangled control objectives, especially for upper body actions.
(ii) \textit{Reward and Postural Stability:} As shown in \Cref{fig:rewards}~(b), \method achieves less tracking errors in both upper-body joints and base angular velocity while in M-WB-RL these two reward terms tend to fluctuate. Additionally, M-WB-RL suffers from larger torso and ankle penalties due to excessive whole-body compensation, resulting in unnatural bending and CoM upright misalignment as shown in \Cref{fig:real_exp}.

\subsection{Real-World Quantatitive Tracking Results}
\begin{figure}[tb]
    \centering
    \begin{minipage}[t]{0.48\linewidth}
        \vspace{0pt}
        \centering
        {\renewcommand{\arraystretch}{2.025}
        \normalsize
        \begin{tabular}{lcc}
            \toprule
            Method & $E_{\text{tracking}}^{\text{upper}}$ & $E_{\text{tracking}}^{\text{root}}$ \\
            \midrule
            Upper-PD-Force-Curr. & 1.81 \tiny$\pm$ 0.13 & \goodnumber{0.40} \tiny$\pm$ 0.04 \\
            M-WB-RL-Force-Curr.       & 0.81 \tiny$\pm$ 0.11 & 0.58 \tiny$\pm$ 0.05 \\
            \goodnumber{\method}   & \goodnumber{0.39} \tiny$\pm$ 0.08 & \goodnumber{0.42} \tiny$\pm$ 0.03 \\
            \bottomrule
        \end{tabular}
        \vspace{0mm}
        \captionof{table}{Real-world Tracking Errors.}
        \label{tab:real-tracking}
        }
    \end{minipage}
    \hfill
    \begin{minipage}[t]{0.48\linewidth}
        \vspace{0pt}
        \centering
        \includegraphics[width=\linewidth]{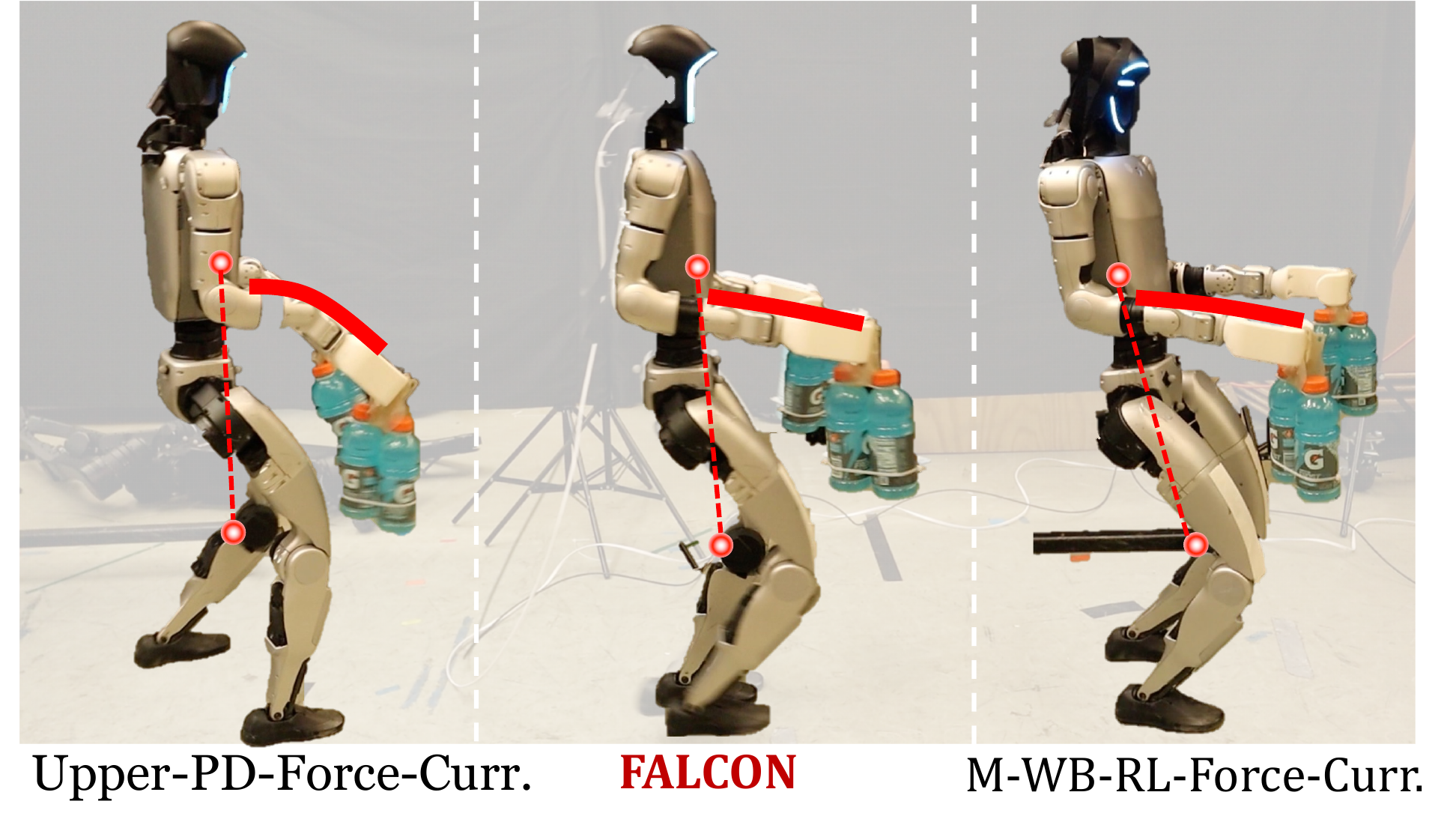}
        \vspace{-6.75mm}
        \captionof{figure}{Real-World Payload Transportation.}
        \label{fig:real_exp}
    \end{minipage}
\vspace{-2mm}
\end{figure}

We evaluate \method on Unitree G1 with each hand loaded with 1.2kg payload in a real-world task, which is walking at $(0.5, 0.0)$m/s with zero angular velocity, fixed height and waist, and keeping the upper body in its default position. We compare against two baselines: (i) \textit{Upper-PD with Force Curriculum}, and (ii) \textit{Monolithic-WB-RL with Force Curriculum}. As shown in \Cref{tab:real-tracking}, \method achieves the lowest tracking errors, and perform stable and natural motion in heavy-duty loco-manipulation.

\subsection{Real-World Deployment with Teleoperation}
To answer \textbf{Q3} (\textit{Does \method work for different humanoids to show cross-platform generalizability? }), we deploy policies trained in simulation on the Unitree G1 and Booster T1 humanoids \textit{\textbf{without} any reward or force curriculum modifications}, thanks to \method efficient dual-agent training and torque-limit-aware 3D force curriculum. As shown in \Cref{fig:teaser}, we evaluate the policies on three forceful loco-manipulation tasks: (1) \textbf{Transporting Payloads}, with 0-20N vertical forces while maintaining stable locomotion and precise upper-body joint tracking; (2) \textbf{Cart-Pulling}, with up to 100N longitudinal (X-Y) forces during walking; and (3) \textbf{Door-Opening}, with up to 40N 3D forces during stance. These force ranges are measured through a force gauge shown in \Cref{appendix:gauge}.

These results demonstrate that \method enables robust policy transfer across platforms with different morphologies and actuation. The learned policies exhibit effective whole-body compensation: the upper body responds adaptively to 3D forces, the lower body leans against significant longitudinal forces, and the base height remains stable under vertical loads.


\subsection{Real-World Deployment with Autonomy}
\begin{figure}
    \centering
    \includegraphics[width=1.0\linewidth]{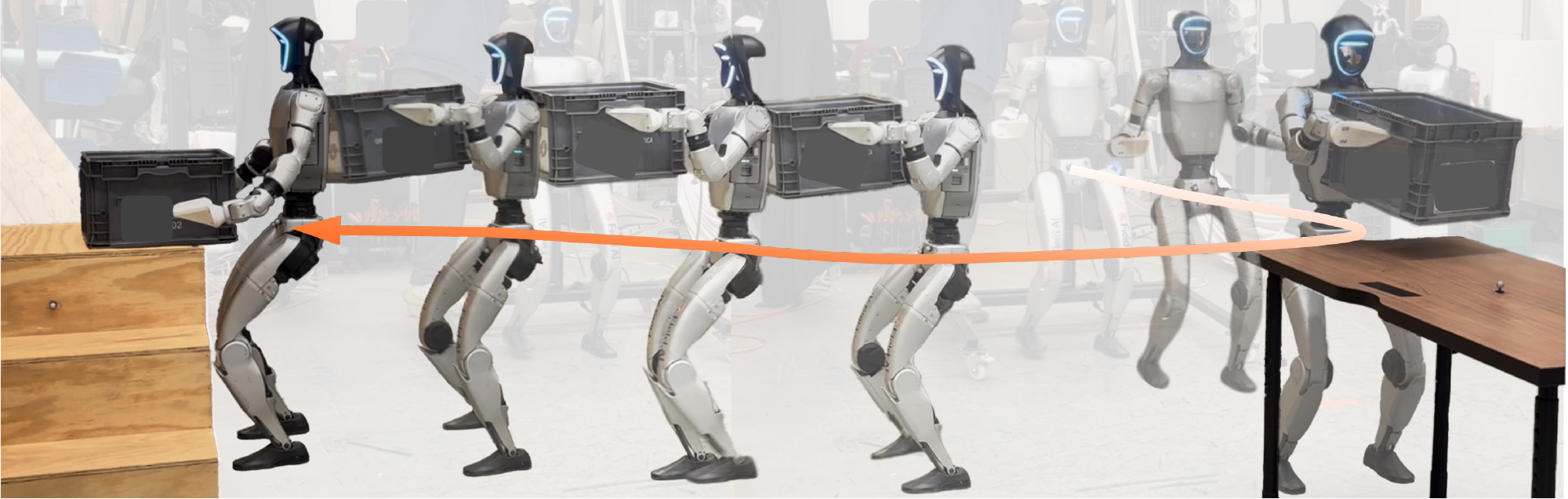}
    \caption{\textbf{Autonomous} Tote Logistics: a humanoid robot walks without a tote, picks up the tote, walks with the tote, and drops off the tote.}
    \label{fig:autonomy}
    \vspace{-4mm}
\end{figure}

We also deploy \method on the Unitree G1 for autonomous tote logistics, a representative warehouse task. As illustrated in \Cref{fig:autonomy}, the robot is required to walk from an initial location to a pickup station, lift a tote of unknown weight, and transport it to a designated area for precise placement. The detailed implementation of the autonomous pipeline can be found in \Cref{appendix:auto}.

\section{Conclusion}
\label{sec:conclusion}
In this paper, we introduce \method, a dual-agent reinforcement learning framework designed for force-adaptive humanoid loco-manipulation. By decoupling the learning of the upper and lower body, while maintaining coordination through shared proprioceptive feedback, \method achieves superior adaptability in handling 3D end-effector forces during complex tasks. Our extensive evaluation demonstrates that \method outperforms both \decouple and \wbc baselines, achieving faster training convergence, reduced tracking errors, and more stable performance across a variety of force regimes. Moreover, \method exhibits strong cross-platform generalizability, successfully transferring policies from simulation to physical humanoids, including tasks like transporting payloads, cart-pulling, and door-opening. These results underscore \method's potential for real-world deployment in forceful interaction scenarios.


\section{Limitations}
\label{sec:limitations}
Despite its strong performance, \method has two key limitations. First, it focuses solely on force disturbances applied to the end-effectors, without accounting for contact forces on other body parts or supporting multi-contact interactions. This restricts its applicability in scenarios involving whole-body support, such as leaning, bracing, or collaborative lifting. Second, the current force curriculum only considers external forces and ignores external torques. As a result, \method may struggle in tasks that involve rotational disturbances, such as operating handles or tools with eccentric loading. Addressing these limitations by incorporating multi-contact reasoning and torque-adaptive policies remains an important avenue for future research.

\section*{Acknowledgments}
We would like to thank our other CMU MRSD Capstone teammates, Ishita Gupta and Shivang Vijay, for their valuable contributions to the autonomy part in this project. We are also grateful to our Capstone advisors, Prof. John Dolan and Prof. Dimitrios Apostolopoulos, for their continuous guidance and support. We acknowledge the hardware support from Unitree Robotics and Booster Robotics. Finally, we thank Haoyang Weng, Wenli Xiao and Yitang Li for their insightful discussions that helped shape the direction of this work.



\bibliography{references}

\clearpage

\appendix
\section{Appendix}
\subsection{Reward Terms}
\label{appendix:reward}
We adopt the similar reward terms from~\cite{he2024omnih2o, cheng2024expressive}, but introduce some important penalties to ensure the locomotion stability under significant external forces, and other tracking rewards for squat and waist twist. The additioanl reward terms are summerized in \Cref{tab:reward}:
\begin{table}[H]
\caption{Additional Reward components and weights: penalty rewards for preventing undesired behaviors for sim-to-real transfer, and task rewards to achieve desired loco-manipulation capability.}
\label{tab:reward}
\centering
\begin{tabular}{  c  c  c  }
\hline
Term                  & Expression & Weight    \\ \hline
               &     Penalty       &           \\ \hline
Hip pos                     &   $\lVert \bs{q}^{hip}_{roll, pitch} \rVert$      & -2.5   \\
Negative knee joint          &   $\sum_j \mathds{1} [q_j < q_j^{\min}]$      & -1.0   \\
Stance tap feet & $|(\bs{p}_{left\_foot} - \bs{p}_{right\_foot})_x|$ in base frame & -5.0 \\
Stance root & $|(\bs{p}_{\text{root}} - \text{mid}(\bs{p}_{\text{feet}}))^y|$ & -5.0 \\
Stand still & $\mathds{1}[\text{no contact}]$ & -0.15 \\
Ankle roll & $\sum_j |q_j^{\text{roll}}|$ & -2.0 \\
\hline
                  &    Task Reward        &           \\ \hline
Root linear velocity x         &    $\exp (-4.0\lVert \bs{v}^x_{t}-\bs{v}^{x*}_{t} \rVert_2) $        & 2     \\
Root linear velocity y         &    $\exp (-4.0\lVert \bs{v}^x_{t}-\bs{v}^{x*}_{t} \rVert_2) $        & 1.5     \\
Root angular velocity &   $\exp (-4.0\lVert \bs{v}^{ang}_{t}-\bs{v}^{ang*}_{t} \rVert_2) $         & 4     \\ 
Root walk height & $\exp\left(-\frac{|\text{command}_z - p_z^{\text{root}}|}{0.05}\right)$ & 2 \\
Waist dofs & $\exp\left(-\frac{\sum_{\theta \in \{\text{yaw, roll, pitch}\}} (\theta^{\text{sim}} - \theta^{\text{cmd}})^2}{0.05}\right)$ & 2 \\
Upper body dofs & $\exp\left(-\frac{\lVert \bs{q}_{\text{upper}} - \bs{q}_{\text{ref}} \rVert_2^2}{0.01}\right)$ & 4 \\

\hline
\end{tabular}%

\end{table}

\subsection{Domain Randomization}
\label{appendix:dr}
We apply the following domain randomization terms during training, which are important for successful sim-to-real transfer.
\begin{table}[H]
\caption{Domain randomization terms including dynamics randomization and external perturbation.}
\label{table:dr}
\centering
\begin{tabular}{ c c }
\hline
Term           & Value                              \\ 
\hline
\multicolumn{2}{c}{\textbf{Dynamics Randomization}}  \\ 
\hline
Friction & $\mathcal{U}(0.5, 1.25)$            \\
Link mass    & $\mathcal{U}(0.9, 1.2) \times \text{default} \ \text{kg}$            \\
Base mass    & $\mathcal{U}(-1.0, 3.0) \ \text{kg}$            \\
P Gain             & $\mathcal{U}(0.9, 1.1) \times \text{default}$          \\
D Gain             & $\mathcal{U}(0.9, 1.1)  \times \text{default} $         \\
Control delay      & $\mathcal{U}(0, 20)\text{ms}$           \\ 
\hline
\multicolumn{2}{c}{\textbf{External Perturbation}}  \\ \hline
Push robot         & $\text{interval}=5s$, $v_{xy}=1 \text{m/s}$                   \\ \hline
\end{tabular}
\end{table}

\subsection{\decouple with Force Estimator}
\label{appendix:force_est}
\begin{figure}[h]
    \centering
    \includegraphics[width=1.0\linewidth]{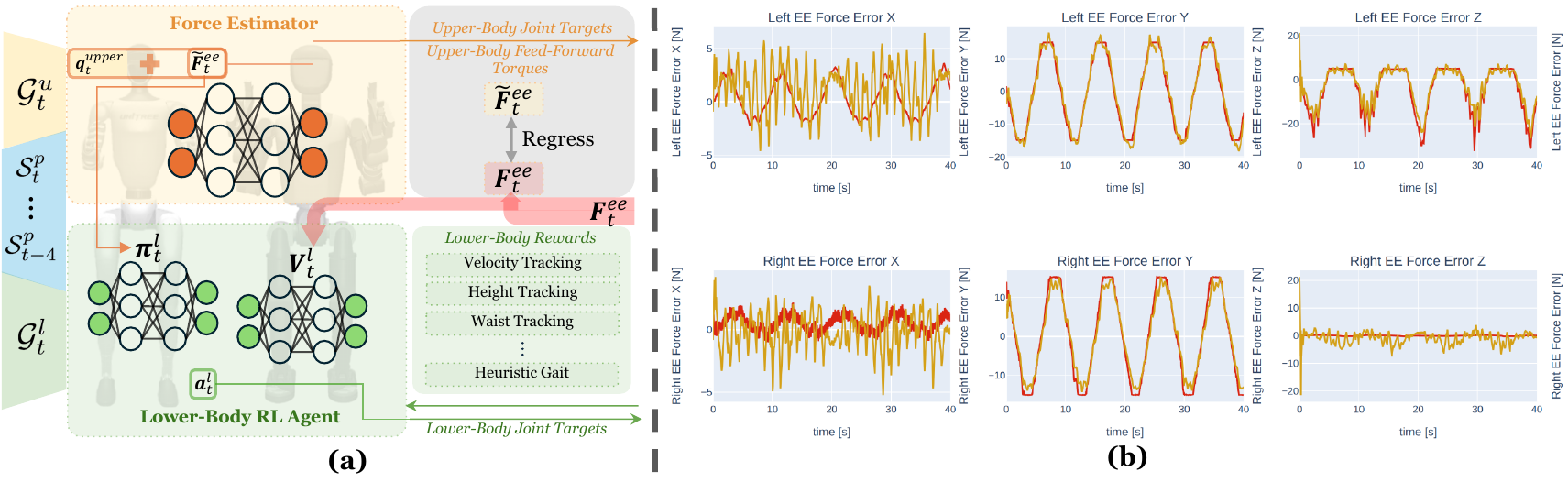}
    \caption{(a) \decouple with Force Estimator; (b) Force Estimator Results: \textcolor{deepyellow}{yellow} lines are the estimated forces while the \textcolor{deepred}{red} lines are the actual forces.}
    \label{fig:force_est}
\end{figure}

We jointly train a 3D force estimator, following a similar approach to~\cite{portela2024learning}, using the robot’s proprioception as input $\bs{s}_t^{\mathrm{p}} \triangleq [\dofposhist, \dofvelhist, \rootangvelhist, \gravityhist, \bs{a}^l_{t-5:t-1}].$
As illustrated in \Cref{fig:force_est}~(a), the estimator predicts the end-effector forces $\Tilde{\bs{F}}_{t}^{ee}$, which are then concatenated with full-body proprioception and fed into the lower-body RL policy. Meanwhile, the upper-body joint torques with force compensation are computed as $\bs{\tau} = K_p (\bs{q}^{\text{upper}}_t - \bs{q}^{\text{upper*}}_t) + K_d \bs{\dot{q}}^{\text{upper}}_t + \bs{J}_{EE}^T \Tilde{\bs{F}}_t^{ee}.$

We compare the estimated and applied forces in \Cref{fig:force_est}~(b), showing close alignment between the two. However, even with accurate force estimates, changes in the contact point on the end-effector alter the effective Jacobian $\bs{J'}_{EE}^T$, making the compensation term $\bs{J}_{EE}^T \Tilde{\bs{F}}_t^{ee}$ inaccurate. Therefore, a force sensor is still necessary during deployment to localize the force application and compute the correct $\bs{J'}_{EE}^T$. Moreover, the compensation assumes quasi-static conditions, introducing additional error when upper-body joints are moving.

\subsection{Force Measurement}
Here, we use Mxmoonfree-Digital-500N-Force-Gauge to measure the peak forces needed for the following force-adaptive tasks: (1) \textbf{Cart-Pulling} for Booster T1 with a Unitree G1 and a Unitree H1 in the cart; (2) \textbf{Door-Opening}; (3) \textbf{Stance-Pulling} for the Unitree G1 and Booster T1.

Here, \textbf{Stance-Pulling} refers to applying longitudinal forces along the X-Y plane while the robot maintains a static stance, and measuring the maximum force it can resist without losing balance. Notably, the Booster T1 demonstrates a higher peak resistive force compared to the Booster T1, primarily due to its lower center of mass (CoM), which contributes to better stability with longitudinal resistance.
\label{appendix:gauge}
\begin{figure}[htb]
    \centering
    \begin{subfigure}[t]{0.49\textwidth}
        \centering
        \includegraphics[width=\textwidth]{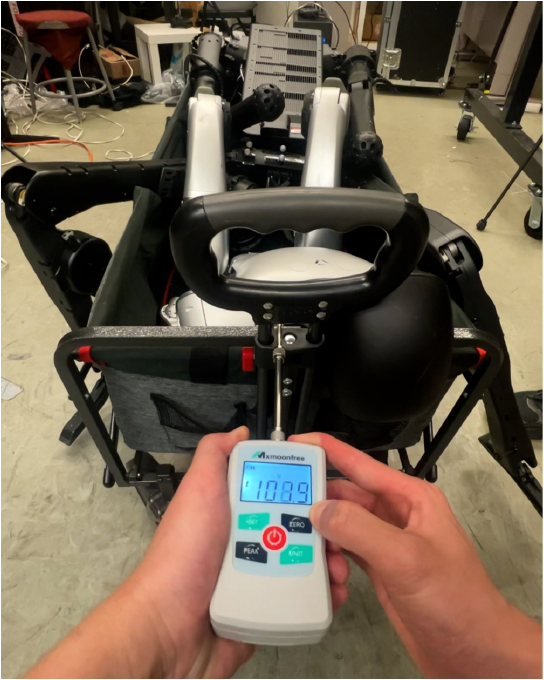}
        \caption{Cart-Pulling (peak: 107.9 N)}
        \label{fig:open_door}
    \end{subfigure}
    \hfill
    \begin{subfigure}[t]{0.49\textwidth}
        \centering
        \includegraphics[width=\textwidth]{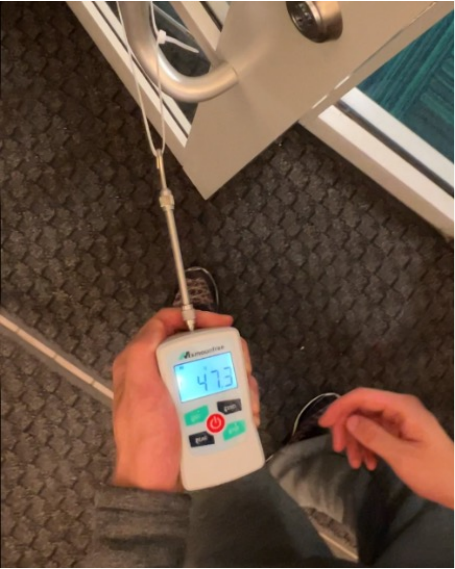}
        \caption{Door-Opening (peak: 47.3 N)}
        \label{fig:g1_push}
    \end{subfigure}

    \vspace{0.4em} 

    \begin{subfigure}[t]{0.49\textwidth}
        \centering
        \includegraphics[width=\textwidth]{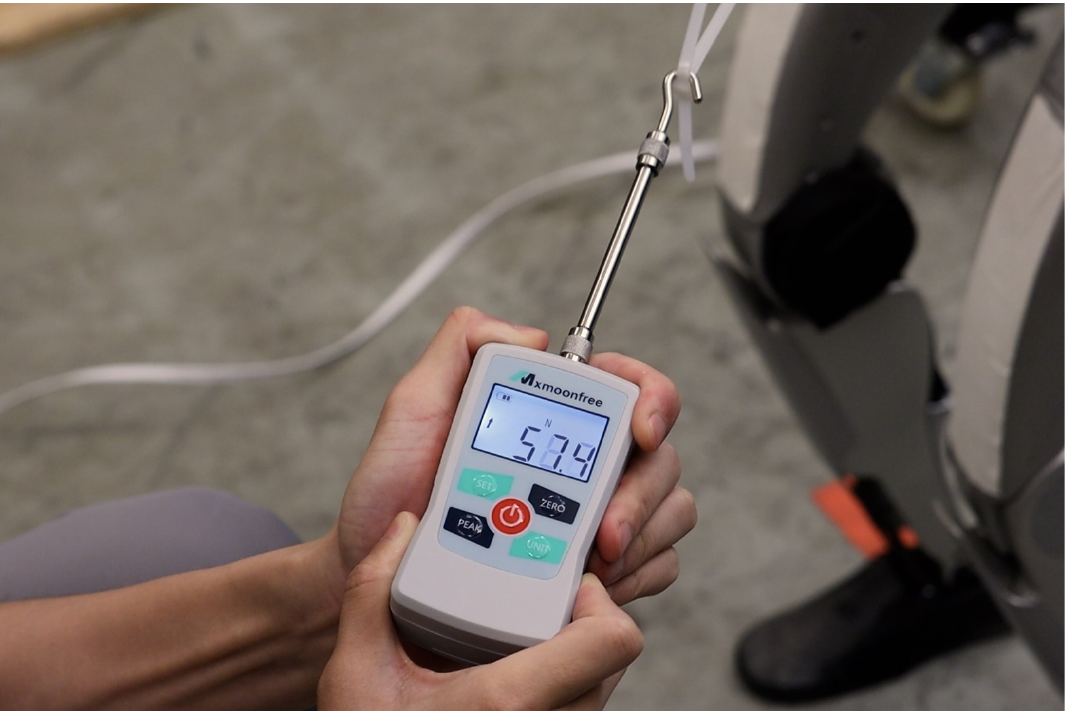}
        \caption{Stance-Pulling: Unitree G1 (peak: 57.4 N)}
        \label{fig:pull_cart}
    \end{subfigure}
    \hfill
    \begin{subfigure}[t]{0.49\textwidth}
        \centering
        \includegraphics[width=\textwidth]{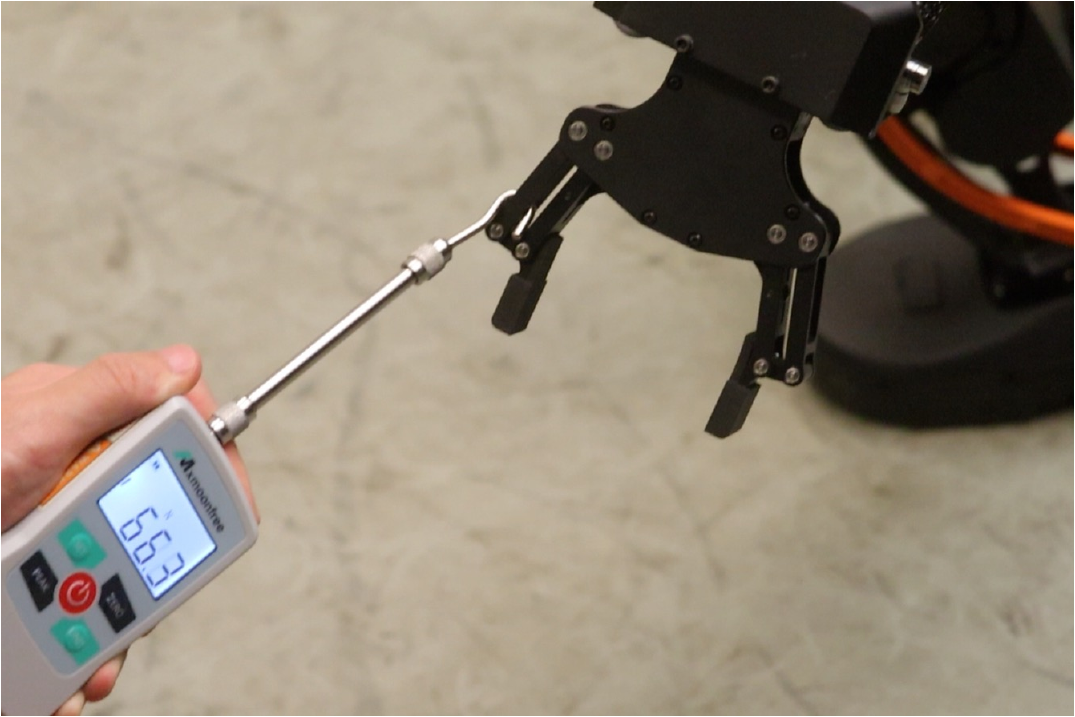}
        \caption{Stance-Pulling: Booster T1 (peak: 66.3 N)}
        \label{fig:t1_gripper}
    \end{subfigure}

    \caption{Maximum force readings captured during different force-adaptive tasks using a handheld force gauge. Subfigures (a)–(d) show peak force values during individual tasks.}
    \label{fig:appendix_force_tests}
\end{figure}

\subsection{Autonomy Pipeline}
\label{appendix:auto}
\begin{figure}[H]
    \centering
    \includegraphics[width=1\linewidth]{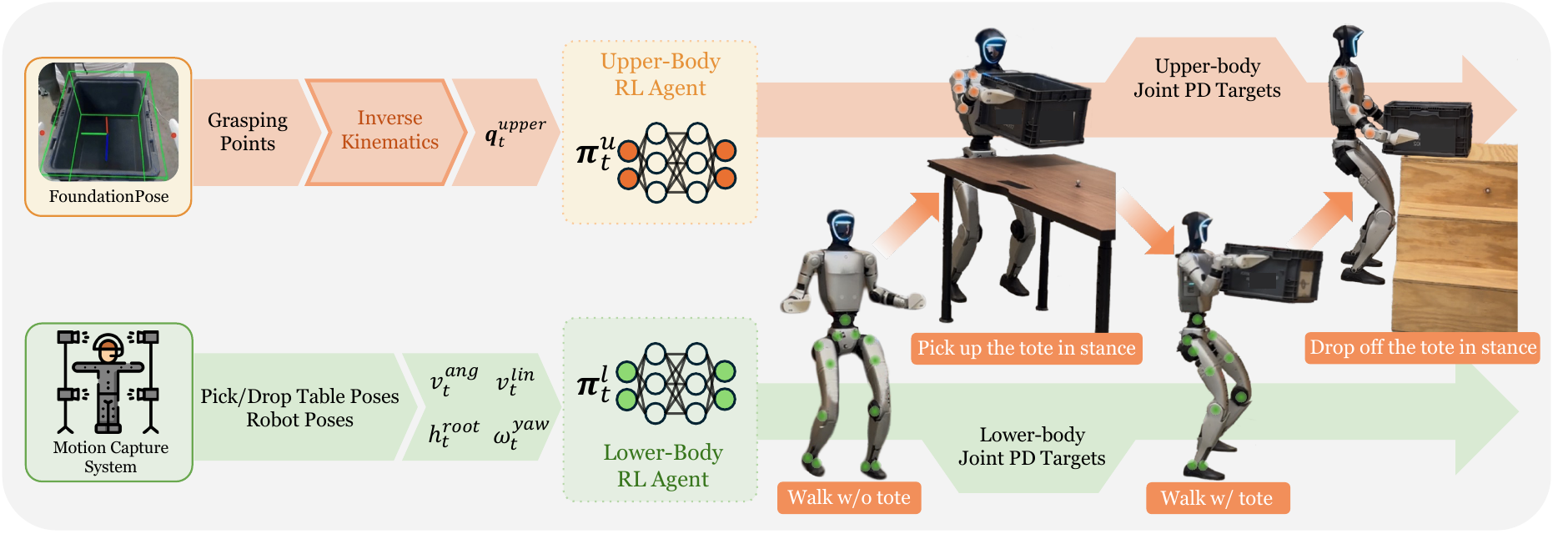}
    \caption{Overview of the autonomy pipeline for \method. The system integrates \method with 6-DoF object pose estimation via FoundationPose, MoCap-based global localization, and inverse kinematics for grasp planning to enable a humanoid robot to perform tote logistics tasks: walk without a tote, pick up the tote, walk with the tote, and drop off the tote.}
    \label{fig:autonomy_pipeline}
\end{figure}

We develop an hierarchically autonomous pipeline for tote logistics, leveraging a Motion Capture (MoCap) system to localize positions of the robot and desks. The robot is controlled by a state-machine framework with four states: (1) walking without the tote, (2) picking up the tote in stance, (3) walking with the tote, and (4) dropping off the tote in stance, as illustrated in Fig.\ref{fig:autonomy_pipeline}. To estimate the tote's pose relative to the camera, we use FoundationPose~\cite{wen2024foundationpose}, a state-of-the-art method for accurate and reliable 6-DoF pose estimation.

\subsubsection{Perception - Pose Estimation}

\begin{figure}[htb]
    \centering
    \includegraphics[width=1\linewidth]{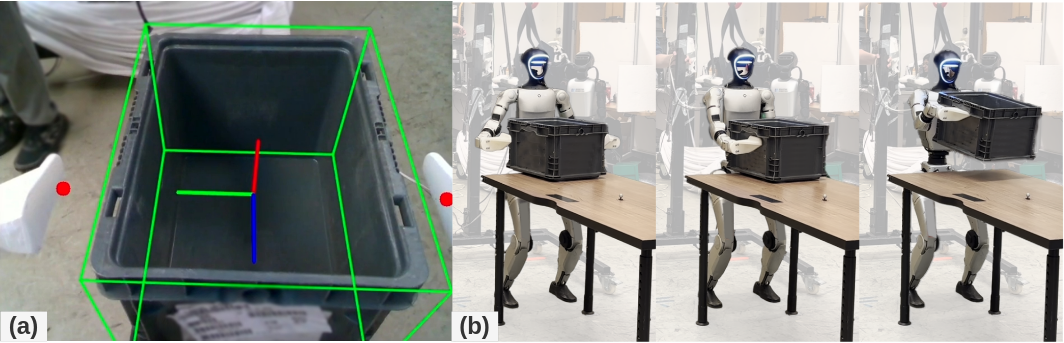}
    \caption{Humanoid G1 Tote Logistics (a) First person view of tote pose estimation (grasping points shown in red). (b) Sequence of actions from left to right- approach, grasp, pickup.}
    \label{fig:grippingpoints}
    \vspace{-2mm}
\end{figure}

To set up the FoundationPose pipeline~\cite{wen2024foundationpose}, we first acquire a high-fidelity 3D model of the industrial tote by performing a raw 3D scan, followed by manual post-processing in a 3D modeling tool. The resulting texture and .obj files serve as inputs to FoundationPose, enabling 6-DoF pose estimation of the tote from the G1 robot's image stream. Additionally, we predefine grasp points longitudinally on the tote's surfaces (Fig.~\ref{fig:grippingpoints}), which are transformed into the robot base frame using the calibrated extrinsics between the camera and the robot.

\subsubsection{Motion Capture System}

The motion capture system provides accurate 6-DoF pose estimates—position \((x, y, z)\) and orientation (yaw, pitch, roll)—for the robot base, pickup table, and drop table in a global reference frame, enabling consistent spatial localization across the system.

When the state machine transitions to ``(2) Pick up the tote in stance'' (see Fig.~\ref{fig:autonomy_pipeline}), triggered by mocap feedback, FoundationPose is executed in real time to estimate the tote pose. The predefined grasp points are then passed to an inverse kinematics (IK) solver, formulated as a go-to-pose problem for the upper-body manipulation.




\subsection{Hardware Limits in Real-World Deployment}
\begin{figure}[htb]
    \centering
    \includegraphics[width=1.0\linewidth]{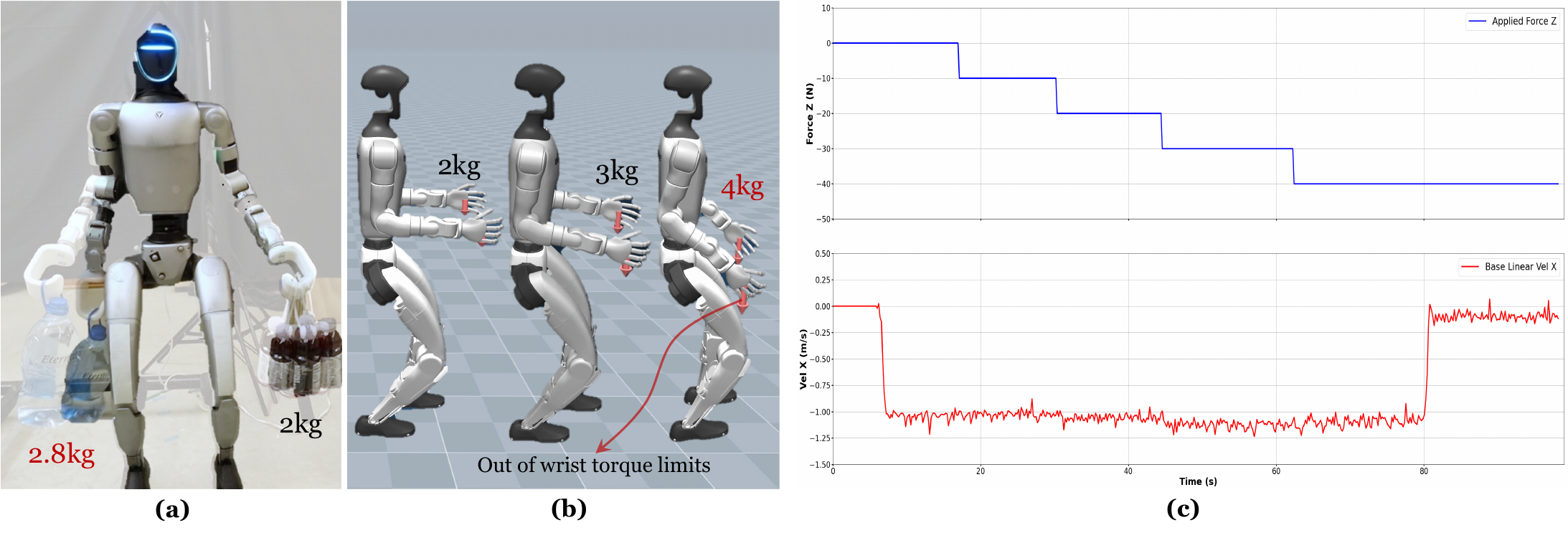}
    \caption{Transporting 0-4kg Payloads in Mujoco}
    \label{fig:mujoco}
\end{figure}

During sim-to-real deployment of the policy trained with \method, we observe that the humanoid robot struggles to sustain high joint torques over extended periods, often leading to rapid motor overheating—particularly at the wrists, as shown in~\Cref{fig:mujoco}~(a). This significantly limits our ability to perform payload transport exceeding 2kg per arm at its default joint position. In contrast, as shown in \Cref{fig:mujoco}~(b) and (c), the same policy evaluated in MuJoCo~\cite{mujoco}, with torque clipping to respect joint limits but without modeling thermal constraints, successfully transports payloads over 3kg per end effector while accurately tracking the linear velocity commanded $-1$ m / s along the x-axis. This highlights a key gap between simulated and real-world actuator endurance.

However, for heavy-duty tasks such as cart-pulling—which require only brief bursts of high torque—the motors are less prone to overheating, as sustained high torque output is not necessary. This enables the robot to successfully perform such tasks in the real world.

\end{document}